\documentclass[letterpaper]{article} 
\usepackage[draft]{aaai2026}  
\usepackage{times}  
\usepackage{helvet}  
\usepackage{courier}  
\usepackage[hyphens]{url}  
\usepackage{graphicx} 
\usepackage{natbib}  
\usepackage{caption} 
\usepackage{algorithm}
\usepackage{algorithmic}

\usepackage{amsmath}
\usepackage{amsfonts}
\usepackage{colortbl}
\usepackage{booktabs} 
\usepackage{xcolor}
\usepackage{subcaption}
\usepackage{subcaption}
\usepackage{multirow}
\usepackage{lipsum} 
\usepackage{multicol} 

\newtheorem{Principle}{Principle}

\usepackage{newfloat}
\usepackage{listings}
\DeclareCaptionStyle{ruled}{labelfont=normalfont,labelsep=colon,strut=off} 
\floatstyle{ruled}
\newfloat{listing}{tb}{lst}{}
\floatname{listing}{Listing}
\title{Continual Adversarial Defense}
\author {
    Qian Wang\textsuperscript{\rm 1}, \
    Hefei Ling\textsuperscript{\rm 1}\thanks{Corresponding author}, \
    Yingwei Li\textsuperscript{\rm 2}, \
    Qihao Liu\textsuperscript{\rm 2}, \
    Ruoxi Jia\textsuperscript{\rm 3}, \
    Ning Yu\textsuperscript{\rm 4}\\
}
\affiliations{
    \textsuperscript{\rm 1}Huazhong University of Science and Technology, China \quad \textsuperscript{\rm 2}Johns Hopkins University, USA\\

    \textsuperscript{\rm 3}Virginia Tech, USA \quad \textsuperscript{\rm 3}Netflix Eyeline, USA\\
    \{yqwq1996, lhefei\}@hust.edu.cn, \quad \{yingwei.li, qliu45\}@jhu.edu, \\
    ruoxijia@vt.edu, \quad   ningyu.hust@gmail.com
}

\usepackage{bibentry}

\begin{document}

\maketitle

\begin{abstract}
In response to the rapidly evolving nature of adversarial attacks against visual classifiers, numerous defenses have been proposed to generalize against as many known attacks as possible.
However, designing a defense method that generalizes to all types of attacks is unrealistic, as the environment in which the defense system operates is dynamic. Over time, new attacks inevitably emerge that exploit the vulnerabilities of existing defenses and bypass them.
Therefore, we propose a continual defense strategy under a practical threat model and, for the first time, introduce the Continual Adversarial Defense (CAD) framework. 
CAD continuously collects adversarial data online and adapts to evolving attack sequences, while adhering to four practical principles:
(1)~continual adaptation to new attacks without catastrophic forgetting,
(2)~few-shot adaptation,
(3)~memory-efficient adaptation, and 
(4)~high classification accuracy on both clean and adversarial data.
We explore and integrate cutting-edge techniques from continual learning, few-shot learning, and ensemble learning to fulfill the principles.
Extensive experiments validate the effectiveness of our approach against multi-stage adversarial attacks and demonstrate significant improvements over a wide range of baseline methods.
We further observe that CAD’s defense performance tends to saturate as the number of attacks increases, indicating its potential as a persistent defense once adapted to a sufficiently diverse set of attacks.
Our research sheds light on a brand-new paradigm for continual defense adaptation against dynamic and evolving attacks.
\end{abstract}


\section{Introduction}
Adversarial attack~\cite{PGD} aims to deceive deep neural networks (DNNs) by adding subtle perturbations to input images, seriously jeopardizing the reliability of DNNs, particularly in domains sensitive to security and trust.
To protect DNNs, researchers have proposed adversarial training (AT)~\cite{TRADES} and purification techniques~\cite{ADP}, which defend against adversarial attacks through one-shot training—where the model enters a static phase after a single defensive stage~\cite{AIR-CAD}. 
However, these approaches often exhibit reduced robustness against adversarial examples and degraded performance on clean images.

\begin{figure}[t]
    \centering
    \includegraphics[width=1\columnwidth]{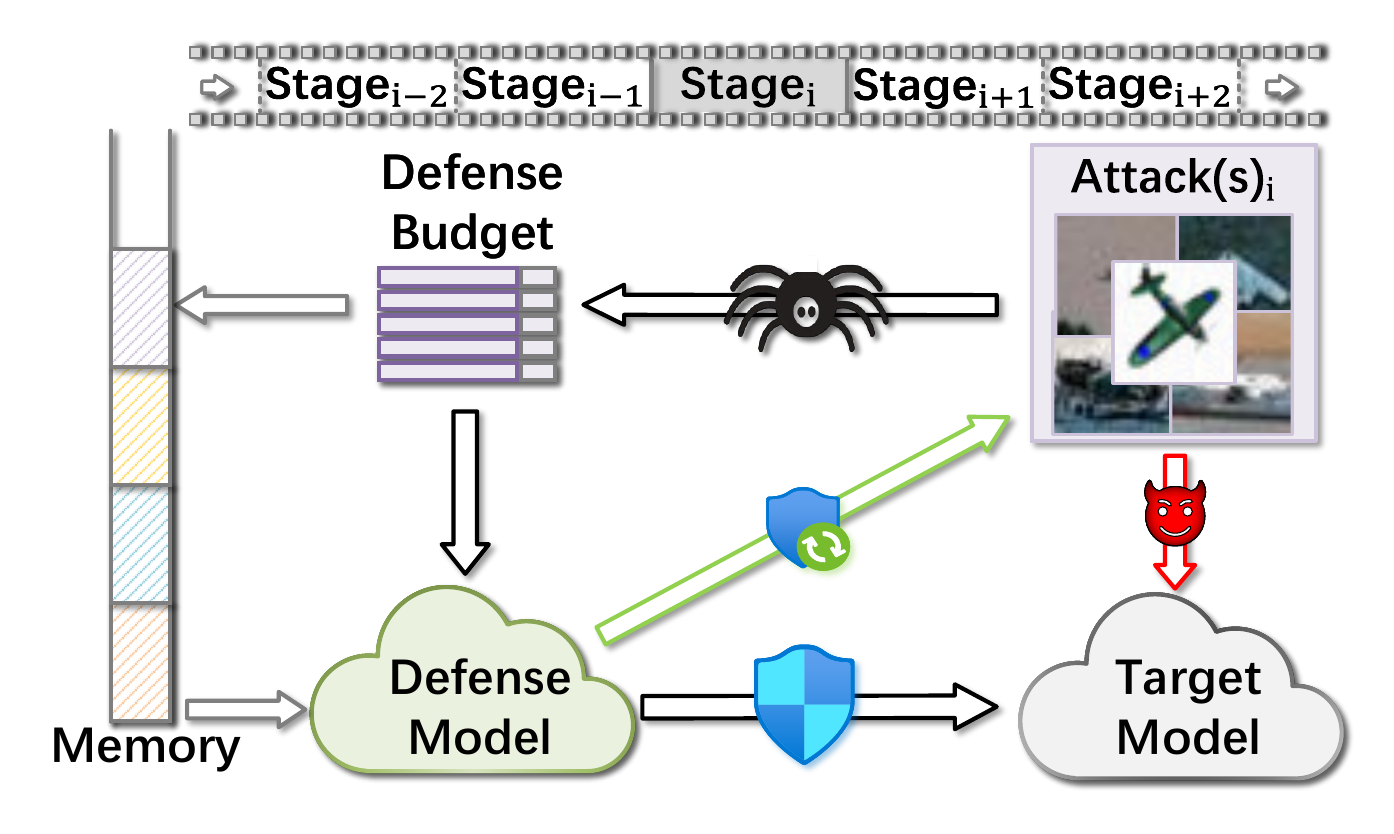}
    \caption{
    In dynamic environments, the defense model protects the target model by continuously collecting a few-shot defense budget and adapting to new attacks, while efficiently managing memory to preserve knowledge of past attacks.
    }
    \label{fig: teaser_figure}
\vspace{-0.5cm}
\end{figure}

Unlike the one-shot defense assumption, real-world defense systems operate in dynamic environments with increasingly aggressive and diverse attacks~\cite{li2023robustness}.
A well-matched approach to the dynamic environments lies in a constantly evolving defense strategy that continuously adapts to newly emerging adversarial attacks~\cite{Test-time-defense}.
However, current model adaptation defenses~\cite{clc-defense} focus on continuously changing defense strategies when under attack rather than leveraging past defense experiences, leading to suboptimal performance.
The concurrent continual defense work~\cite{AIR-CAD} treats adversarial attacks as a task sequence and tries to train a robust model via lifelong learning under a white-box setting.
However, the poor performance on both adversarial and clean images hinders its deployment in real-world scenarios and calls into question the reasonableness of its continual defense setting.

To eliminate the insufficiency of the above defense setting, we put forward a practical defense deployment and a challenging threat model.
As illustrated in Figure~\ref{fig: teaser_figure}, the defense system is deployed in the cloud and continuously adapts to newly emerging attacks at each stage by leveraging a defense budget—comprising adversarial examples and their ground-truth labels—collected from web crawlers, the security department, and users of the target model in real-world scenarios.
In practice, a stage may consist of only a few samples from one or multiple attacks, and a new stage begins once the system accumulates a sufficient defense budget.
When adapting to newly emerging attacks, the defense system is allowed to access the defense budgets from previous stages cached in memory, to preserve knowledge of past attacks.
The defense system also has access to an initial attack and shares the same training data and architecture as the target model. 
Meanwhile, attacks are initiated under a gray-box setting, wherein they possess knowledge of the classifier's architecture and have access to training data, but remain unaware of the defense mechanism implemented.

Taking real-world constraints into account, we propose four principles for continual defense:
(1)~\textit{Continual adaptation to new attacks without catastrophic forgetting.}
As a defender, it is crucial to effectively adapt to a variety of new attacks across different stages while preserving knowledge from previous encounters.
(2)~\textit{Few-shot adaptation.}
An increase in the defense budget indicates a heightened risk of attacks on the target model. Therefore, it is crucial to initiate adaptation before the defense budget grows excessively.
(3)~\textit{Memory-efficient adaptation}.
Over time, the growing influx of attacks leads to the accumulation of the defense budget, which may eventually exceed memory limits, posing practical challenges under constrained capacity.
(4)~\textit{High classification accuracy on both clean and adversarial data}.
A robust defense should not compromise the performance on benign inputs for the sake of adversarial robustness.
Therefore, ensuring high classification accuracy on both clean and adversarial data is paramount.

In this paper, we propose the first Continual Adversarial Defense (CAD) framework, designed to defend against evolving attacks in a stage-wise manner using a few-shot defense budget and efficient memory.
In the first stage, we use adversarial data from an initial attack (e.g., PGD) to train an initial defense model that is specifically designed to classify adversarial examples and complement the target model’s predictions.
Inspired by continual learning (CL), we treat the new classes introduced by each attack as incremental and expand the classifier accordingly, followed by fine-tuning the expanded layer using a few-shot defense budget.
To address the overfitting issue inherent in few-shot fine-tuning~\cite{FACT}, we pre-assign virtual classes to compress the embeddings of past attacks, thereby reserving embedding space in the defense model for future attacks.
To achieve memory efficiency, we employ prototype augmentation~\cite{SSRE}, which enables the preservation of decision boundaries from previous stages without explicitly storing any defense budget.
Simultaneously, a lightweight model is employed to ensemble the defense model and the target model by estimating reliable logits for the input, thereby maintaining classification accuracy on both clean and adversarial data.

Extensive experiments on CIFAR-10 and ImageNet-100 demonstrate that CAD effectively defends against multi-stage attacks under a few-shot defense budget, while maintaining high accuracy on clean data.
To the best of our knowledge, this is the first exploration of a Continual Adversarial Defense and its real-world deployment.\footnote{We can provide timestamps to support this claim.}
In addition, we observed that CAD’s defense performance tends to saturate as the number of attacks increases, indicating its potential to serve as a persistent defense mechanism once adapted to a sufficiently diverse set of attacks.


Our main contributions can be summarized as follows:

$\bullet$ To counter evolving adversarial attacks, we propose a continual defense strategy under a practical threat model, where diverse attacks emerge progressively across stages.
The defense system must adapt to these attacks using a few-shot defense budget and memory-efficient mechanisms, while preserving knowledge of past attacks and maintaining high performance on both clean and adversarial data.



$\bullet$ We propose, for the first time, the Continual Adversarial Defense (CAD) framework, which defends against attacks in dynamic scenarios by adhering to four practical principles: continual adaptation without catastrophic forgetting, few-shot adaptation, memory-efficient adaptation, and high classification accuracy on both clean and adversarial data.
To fulfill these principles, CAD integrates cutting-edge techniques from continual learning, few-shot learning, non-exemplar class-incremental learning, and ensemble learning.


$\bullet$ Extensive experiments validate the effectiveness of CAD in defending against multi-stage adversarial attacks under a few-shot budget and memory-efficient constraints, consistently outperforming baseline methods.
Moreover, we observe that CAD’s defense performance tends to saturate as the number of attacks increases, suggesting its potential to serve as a persistent defense mechanism once adapted to a sufficiently diverse set of attacks.


\section{Related Work}

\subsection{Adversarial Defense}
Researchers have proposed various general defense methods against adversarial attacks~\cite{DIFGSM, NIFGSM}. 
As a standard defense method, adversarial training~\cite{JEM, PAT, RPF, FastAdv} aims to enhance the robustness of the target model by training it with adversarial examples.
Another branch of adversarial defense involves purifying the data stream~\cite{EBM, DnC, DiffPure} to remove potential adversarial perturbations or noise that could deceive the model. 
To counter attacks in dynamic environments, model adaptation methods~\cite{lyapunov-defense, ea-dynamic, ASODE-dynamic} continuously adjust their parameters, state, or activations when encountering attacks to enhance robustness.
In fact, all of the aforementioned methods exhibit limited robustness, leading to poor classification performance on both adversarial and clean data.
Our CAD framework provides an effective solution for countering a wide spectrum of stage-wise emerging attacks, while ensuring robust performance on clean data.

\subsection{Continual Learning}
Continual learning (CL)~\cite{lyy2022online} aims to learn from a sequence of new classes while preserving knowledge of previously learned ones.
Many works have been proposed for CL~\cite{iCarl, Dual_Aug_CIL, PASS}: 
In recent years, some methods have attempted to solve the CL problem without relying on the need to preserve data (non-exemplar)~\cite{SSRE}, and few-shot scenarios~\cite{CEC, FACT} in which only a small number of new class data are available.
Attempting to break the one-shot defense assumption, a concurrent work~\cite{AIR-CAD} is proposed to alleviate catastrophic forgetting in a simple continual defense scenario.
In this paper, we convert the proposed defense scenario into a few-shot and non-exemplar CL setting, which necessitates the defense mechanism to utilize a few-shot defense budget and efficient memory for adaptation.

\section{Threat Model and Defense Principles}
\label{Scenario}

The defense system is deployed in a networked environment to protect the target model against a wide spectrum of evolving attacks in dynamic real-world scenarios.
Over time, new attacks are continuously developed, potentially producing poisoned examples on the Internet to exploit vulnerabilities in the target model and trigger security breaches.
In light of this reality, we propose a practical threat model in which distinct attacks emerge across stages, requiring the defense system to adapt to evolving threats by leveraging a defense budget (i.e., adversarial examples and their ground-truth labels).
\textbf{Typically, a stage may consist of only a few samples from one or multiple attacks, and a new stage begins once the system has accumulated a sufficient defense budget}. 
The defense budget plays a crucial role in shaping the defender’s response to emerging threats. In real-world scenarios, such threats are often identified during screening or cleanup processes, and the defense budget is typically provided by web crawlers, security services, and users of the target model.
We formulate the continual defense scenario as follows.

\begin{figure*}[ht]
    \centering
    \includegraphics[width=2.12\columnwidth]{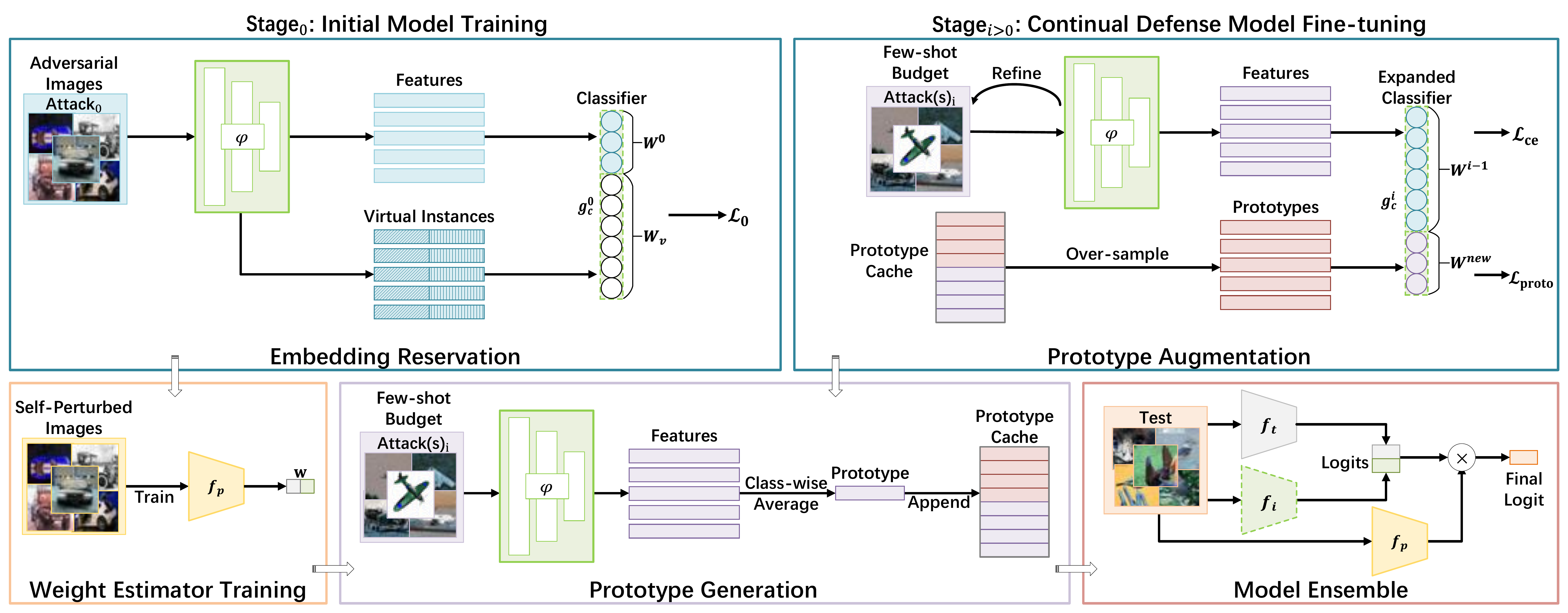}
    \caption{
    Illustration of CAD. 
    Treating new attack classes as incremental, we use continual adaptation to protect the target model while preventing catastrophic forgetting. 
    In stage-0, we reserve embedding space during initial training to mitigate future overfitting. 
    A weight estimator model $f_p $ is trained for the model ensemble. 
    In stage-$i$ ($i = 1,2,\cdots, T$), we freeze the feature extractor $\varphi$ of $f_i$, fine-tune the expanded classifier $g_c^i$ with a few-shot defense budget, and store class-wise prototypes in a cache for memory efficiency. 
    Finally, the target model $f_t$ and the defense model $f_i$ are ensembled via $f_p$ for robust classification.
    }
    \label{fig: pipeline}
\end{figure*}

This scenario is established on the $N$-way $K$-shot classification paradigm.
In the gray-box setting~\cite{Taran:19defending}, the attacker has access to the architecture of the target model ${f}_t: \mathcal{X} \to \mathbb{R}^N$, but not the defense mechanism. 
The target model is trained on $\mathcal{D}_\mathrm{train}$ and evaluated on $\mathcal{D}_\mathrm{test}$, where $\mathcal{X}$ is the image space.
The defender is allowed to use $\mathcal{D}_\mathrm{train}$ and has full knowledge of the initial attack $A_0(\cdot)$ at stage-0.
At stage-$i$ ($i = 1, 2, \cdots, T$), new attacks $A_i(\cdot)$ emerge, and the defender receives a defense budget $\mathcal{A}_\mathrm{train}^i = \{(\mathbf{x}_{\mathrm{adv}}^i,y) | \mathbf{x}_{\mathrm{adv}}^i = A_i(\mathbf{x}), (\mathbf{x},y) \in \mathcal{D}_\mathrm{train}\}$\footnote{In the adversarial attack formula $\mathbf{x}_{\mathrm{adv}}^i = A_i(\mathbf{x},y,f_t)$, we omit the target model $f_t$ and the ground-truth $y$.} consisting of $N \times K$ samples—i.e., $K$ samples per class for $N$ classes—which is used to adapt to the new attack. Here, $y$ denotes the ground-truth label of sample $\mathbf{x}$.
Evaluations are conducted on $\mathcal{D}_\mathrm{test}$ and $\{\mathcal{A}_\mathrm{test}^k\}_{k=0,1,\dots,i}$ at each stage-$i$, where $\mathcal{A}_\mathrm{test}^i = \{(\mathbf{x}_{\mathrm{adv}}^i, y) | \mathbf{x}_{\mathrm{adv}}^i = A_i(\mathbf{x}), (\mathbf{x},y) \in \mathcal{D}_\mathrm{test}\}$.

Taking reality into account, the defense mechanism should satisfy the following principles:

\begin{Principle}
\label{principle: adaptivity}
Continual adaptation to new attacks without catastrophic forgetting.
\end{Principle}
The defense mechanism must be capable of adapting to a range of novel attacks that emerge over time, while preserving knowledge of previously encountered ones.
\begin{Principle}
\label{principle: few-shot}
Few-shot adaptation.
\end{Principle}
An increase in the defense budget directly correlates with the frequency of successful attacks on the target model.
To prevent potential security breaches caused by delayed responses, the defense system must proactively adapt using a few-shot budget.
\begin{Principle}
\label{principle: limited-memory}
Memory-efficient adaptation.
\end{Principle}
Over time, the continuous influx of attacks leads to the accumulation of defense budgets, potentially causing memory constraints.
In practice, the defender may lack sufficient memory capacity to store all received data.
Therefore, retaining the entire defense budget for adaptation is impractical and should be avoided.
\begin{Principle}
\label{principle: acc}
High classification accuracy on both clean and adversarial data
\end{Principle}
The previous defense strategies have led to a sacrifice in the classification accuracy of clean data, resulting in performance degradation. 
Such declines can adversely impact critical real-world applications.
As a result, it is imperative for a defender to maintain performance on both clean and adversarial data.

\section{Continual Adversarial Defense Framework}

As shown in Figure~\ref{fig: pipeline}, the Continual Adversarial Defense (CAD) framework comprises four key components:
(1) continual adaptation to defend the target model while mitigating catastrophic forgetting;
(2) embedding reservation to alleviate overfitting caused by the few-shot adaptation;
(3) prototype augmentation to achieve memory efficiency; and
(4) model ensemble to ensure high classification accuracy on both clean and adversarial data.

\subsection{Continual Adaptation to New Attacks}
\label{layer expansion}
In response to Principle~\ref{principle: adaptivity}, we employ continual adaptation to defend against new attacks. 
In the beginning (stage-0), an initial defense model $f_0: \mathcal{X} \to \mathbb{R}^N$ which consists of a feature extractor $\varphi: \mathcal{X} \to \mathbb{R}^d$ and a classifier $g_c^{0}: \mathbb{R}^d \to \mathbb{R}^N$ is optimized under full supervision using an adversarial dataset $\mathcal{A}_\mathrm{train}^0 = \{(\mathbf{x}_{\mathrm{adv}}^0,y)|\mathbf{x}_{\mathrm{adv}}^0 = A_0(\mathbf{x}), (\mathbf{x},y) \in \mathcal{D}_\mathrm{train}\}$.
The defense model is designed to tackle adversarial examples, complementing the target model.
At stage-$i$ ($i = 1, 2, \cdots, T$), the defense model ${f}_i: \mathcal{X} \to \mathbb{R}^N$ adapts to new attacks using the defense budget $\mathcal{A}_\mathrm{train}^i$, while mitigating catastrophic forgetting—required by Principle~\ref{principle: adaptivity}—through cached prototypes $P_e$ extracted from past attacks.

Inspired by continual learning (CL)~\cite{lyy2022online}, we treat the new classes introduced by each attack as incremental classes and expand the classifier from $g_c^{i-1}:  \mathbb{R}^d \to \mathbb{R}^{N\times i}$ to $g_c^{i}: \mathbb{R}^d \to \mathbb{R}^{N\times {(i+1})}$ at stage-$i$ ($i = 1, 2, \cdots, T$).
The parameters of the expanded classifier consist of those inherited from the previous classifier and newly initialized weights: $W^i = [W^{i-1}, W^{\text{new}}]$, where $W^i = [\mathbf{w}^i_1, \cdots, \mathbf{w}^i_{N_i}]$ denotes the parameter matrix of $g_c^i$.
Meanwhile, the ground-truth label of each adversarial example $\mathbf{x}_{\mathrm{adv}}^i \in \mathcal{A}_\mathrm{train}^i$ is reassigned to correspond to the incremental classes introduced by the new attack \footnote{We omit the networks' parameter in formulas.}:
\begin{equation}
\label{eq:rewrite_y}
    y^i = y + N \times {i}.
\end{equation}

Due to the intrinsic similarity among certain adversarial attacks~\cite{Wang:23selfperturbation}, a defense model trained on one attack can partially generalize to others, potentially rendering some class expansions unnecessary.
Past attacks recurring in new stages also lead to unnecessary adaptation.
Therefore, before fine-tuning at each stage, we test the model to filter out redundant data in the defense budget—that is, samples that are already correctly classified.

For evaluation, we focus solely on the class assignment of the image, rather than the specific attack to which it is subjected.
Therefore, the prediction for an instance $\mathbf{x}$ is:
\begin{equation}
    y_{\mathrm{pred}} = (\arg\max f_i(\mathbf{x})) \bmod{N}.
\end{equation}

In the following section, we detail the training of the initial defense model $f_0$. 
After training, we freeze the feature extractor $\varphi$ and fine-tune the expanded portion of the classifier for each new attack $A_i$ using the few-shot budget $\mathcal{A}_{\mathrm{train}}^i$.

\subsection{Embedding Space Reservation for Few-Shot Adaptation}
\label{sec: embedding space reserving}
A problem brought by the few-shot budget in Principle~\ref{principle: few-shot} is overfitting, which is a well-known issue in few-shot learning~\cite{protonet}.
To mitigate this issue, we pre-assign virtual classes to compress the embeddings of past attacks, thereby reserving embedding space in the defense model for future attacks.

First, before training the initial defense model $f_0$, several virtual classes $W_v = [\mathbf{w}_v^1, \cdots , \mathbf{w}_v^V] \in \mathbb{R}^{d\times V}$ are pre-assigned in the classifier~\cite{CEC}, where $V = N \times T$ is the number of virtual classes, i.e., the reserved classes for future attacks. 
Therefore the output of the current defense model is $f_0(\mathbf{x}) = [W^0, W_v]^{\top} \varphi(\mathbf{x})$. 
After training, $W_v$ will be used to initiate the new parameters of the expanded classifier.

Second, virtual instances are constructed by manifold mixup~\cite{manifoldmix}:
\begin{equation}
    \mathbf{z} = h_2 ( \omega h_1(\mathbf{x}_{\mathrm{adv},r}^0) + (1-\omega)h_1(\mathbf{x}_{\mathrm{adv},s}^0) ), 
\end{equation}
where $\mathbf{x}_{\mathrm{adv},r}$ and $\mathbf{x}_{\mathrm{adv},s}$ belong to different classes $r$ and $s$, and $\omega \sim \mathrm{B}(\alpha, \beta)$ is a trade-off parameter the same as~\cite{FACT}. 
$h_1$ and $h_2$ are decoupled hidden layers of feature extractor i.e. $\varphi(\mathbf{x}) = h_2(h_1(\mathbf{x}))$.

Finally, the embedding space reservation is conducted by training $f_0$ with the following loss function:
\begin{align}
\label{eq:space_reserve}
     \mathcal{L}_{0} = &F_{\mathrm{ce}}(f_0(\mathbf{x}_\mathrm{adv}^0), y^0) +  \gamma F_{\mathrm{ce}}(\mathsf{Mask}(f_0(\mathbf{x}_{\mathrm{adv}}^0), y^0), \hat{y}) + \nonumber\\
      + &F_{\mathrm{ce}}(f_0(\mathbf{z}), \hat{y})+ \gamma F_{\mathrm{ce}}(\mathsf{Mask}(f_0(\mathbf{z}), \hat{y}), \hat{\hat{y}}),
\end{align}
where $\hat{y} = \mathrm{argmax}_j \mathbf{w}_v^{j\top} \varphi(\mathbf{x}_{\mathrm{adv}}^0) + N$ is the virtual class with maximum logit, acting as the pseudo label. 
$\hat{\hat{y}} = \mathrm{argmax}_k \mathbf{w}_k^{0\top} \mathbf{z}$ is the pseudo label among current known classes.
$\gamma$ is a trade-off parameter, $F_{\mathrm{ce}}$ represents the standard cross-entropy loss~\cite{CrossEntropyLoss}, and function $\mathsf{Mask}(\cdot)$ masks out the logit corresponding to the ground-truth:
\begin{equation}
\label{eq:Mask_f0}
    \mathsf{Mask}(f_0(\mathbf{x}_{\mathrm{adv}}^0),y^0) = f_0(\mathbf{x}_{\mathrm{adv}}^0) \otimes (\mathbf{1}- \mathrm{onehot}(y^0)),
\end{equation}
where $\otimes$ is Hadamard product and $\mathbf{1}$ is an all-ones vector.

In Eq.~\ref{eq:space_reserve}, pushes the adversarial instance $\mathbf{x}_{\mathrm{adv}}^0$ towards its ground truth (item 1) and away from the reserved virtual classes (item 2).
Simultaneously, it drives the virtual instance $\mathbf{z}$ towards the corresponding virtual classes (item 3) and away from other classes (item 4).
Trained with $\mathcal{L}_{0}$, the embeddings of the initial benign classes become more compact, and the embedding space for virtual classes is reserved~\cite{FACT}. 
The reserved space allows the defense model to be adapted more easily in the future and alleviates overfitting brought by the few-shot budget.

\subsection{Prototype Augmentation for Memory Efficiency}
\label{sec:prototype augmentation}
In response to Principle~\ref{principle: limited-memory}, we use prototype augmentation to achieve memory efficiency.


When learning new classes, the decision boundaries of previously learned classes can change significantly, resulting in a severely biased unified classifier~\cite{PASS}.
To mitigate this issue, many CL methods store a subset of past data and jointly train the model with both past and current data.
However, preserving the past defense budget may also lead to memory shortage.
To achieve memory efficiency, we adopt prototype augmentation~\cite{SSRE} to maintain the decision boundary of previous stages, without saving any defense budget.
We store one prototype in the deep feature space for each class under each attack, and over-sample (i.e., $U_{B}$) the prototypes $\mathbf{p}_B$ and their corresponding ground-truth labels $y_B$ to the batch size, thereby calibrating the classifier.
\begin{equation}
\label{eq:loss proto}
    \mathbf{p}_B = U_{B}(P_e), \quad \mathcal{L}_{\mathrm{proto}} = F_{\mathrm{ce}}(\mathbf{p}_B, y_B),
\end{equation}
where $P_e=\{\mathbf{p}_e^j\}_{j=0}^{N\times i}$ denotes the prototype cache, i.e. the set of class-wise average embeddings~\cite{protonet}, which is defined as:
\begin{equation}
\label{eq:prototype}
    \mathbf{p}_{e}^j = \frac{1}{K} \textstyle{\sum_{k=1}^{|\mathcal{A}_{\mathrm{train}}^i|}} \mathbb{I} (y_{k}^i=j) \varphi(\mathbf{x}_{\mathrm{adv}, k}^i),
\end{equation}
where $\mathbb{I}(\cdot)$ is the indicator function, $K$ the amount of budget for each class, and $(\mathbf{x}_{\mathrm{adv}, k}^i, y_{k}^i)$ the $k$-th sample in $\mathcal{A}_{\mathrm{train}}^i$.

After each adaptation, prototypes of the current attack are added to cache $P_e$.
At each stage, the defense model is adapted using the following loss function:
\begin{equation}
\label{eq:finetune}
    \mathcal{L}_f = \mathcal{L}_{\mathrm{ce}} + \lambda \mathcal{L}_{\mathrm{proto}}, \ \textrm{where} \ \mathcal{L}_{\mathrm{ce}} = F_{\mathrm{ce}}(f_i(\mathbf{x}_\mathrm{adv}^i), y^i).
\end{equation}

\subsection{Model Ensemble for Clean Data Classification}
\label{sec: Ensemble}

To satisfy Principle~\ref{principle: acc}, which requires high classification accuracy on both clean and adversarial data, we introduce a model ensemble as the final component of CAD.
Ensemble adversarial training (EAT)~\cite{Ensemble_AT} is a simple yet effective method for defending against adversarial attacks in a gray-box setting, by training a robust model using adversarial examples generated by the target model, while maintaining classification performance on clean data.
We extend EAT to our scenario by introducing a lightweight weight estimator $f_p$ that adaptively fuses the logits of the defense model and the target model.

We adopt self-perturbation~\cite{Wang:23selfperturbation} to train the weight estimator $f_p$ at the first stage.
When trained with self-perturbation, the weight estimator becomes agnostic to specific attacks and can effectively distinguish between clean and adversarial images.
The weight estimator $f_p$ outputs a weight vector $\mathbf{w}\in \mathbb{R}^2$ to ensemble the defense model and the target model:
\begin{equation}
\label{eq:model_ensmeble}
    \mathrm{logit}_i(\mathbf{x}) =  \mathbf{w} \cdot [f_t(\mathbf{x}), f_i(\mathbf{x})]^{\top}, \ \mathrm{where} \ \mathbf{w} = f_p(\mathbf{x}).
\end{equation}
In this way, both clean and adversarial data are routed to their corresponding models, increasing the likelihood of correct classification.

For the overall algorithm of the CAD framework and the details about self-perturbation, please refer to the supplementary materials.

\section{Experiment}


To evaluate the performance of CAD against various adversarial attacks in a dynamic scenario, we conduct extensive experiments and compare CAD with baselines from four research streams.
For evaluation and analysis, we employ two metrics in this section:
(1) Classification accuracy (Acc) after each adaptation step, used to assess defense performance.
(2) Average Incremental Accuracy (AIAcc)~\cite{iCarl}, defined as the average classification accuracy over all attacks encountered up to the current stage, used to evaluate the model’s ability to mitigate catastrophic forgetting.

\subsection{Experiment Settings}
In this section, we introduce the datasets, adversarial attack configurations, defense baselines, and implementation details. Additional implementation details can be found in the supplementary materials.

\subsubsection{Datasets}

Experiments are conducted on CIFAR-10 and ImageNet-100.
CIFAR-10~\cite{CIFAR10} consists of 10 classes, each containing 5,000 training images and 1,000 test images.
ImageNet-100~\cite{ImageNet100} is a subset of ImageNet~\cite{ImageNet}, comprising 100 classes with 1,000 training images and 100 test images per class.
Images from CIFAR-10 and ImageNet-100 are resized to $32 \times 32$ and $224 \times 224$, respectively, and undergo data augmentation including horizontal flipping and random cropping.

\subsubsection{Attacks}
We choose PGD-$\ell_{\infty}$~\cite{PGD} as the initial attack.
After that, 8 adversarial attack methods under $\ell_{\infty}$: SNIM~\cite{NIFGSM}, BIM~\cite{BIM}, RFGSM~\cite{Ensemble_AT}, MIM~\cite{MIFGSM}, DIM \cite{DIFGSM}, NIM~\cite{NIFGSM}, VNIM and VMIM~\cite{VNIFGSM} compose the attack pool in defense performance evaluation.
To evaluate CAD's effectiveness in mitigating catastrophic forgetting, $\ell_{\infty}$ attacks (VMIM and SNIM), $\ell_2$ attacks (CW and DeepFool), $\ell_1$ attacks (EAD and EADEN), and $\ell_0$ attacks (OnePixel and SparseFool) compose the attack pool.

\begin{table*}[ht]
\footnotesize
    \centering
    \tabcolsep=0.137cm
    \caption{
    Classification accuracy (Acc) against $\ell_{\infty}$ attacks at each stage on ImageNet-100. The column “Clean” denotes the standard accuracy on clean images. 
    “AT” and “AP” denote adversarial training and purification baselines, respectively.
    $\mathrm{\textbf{CAD}}^\dagger$ is Premium-CAD without memory or defense budget constraints, serving as the empirical upper bound (gray background).
    }
    \begin{tabular}{c|l|ccccccccc|c}
        \toprule
        \multicolumn{2}{c|}{Method} & 0:PGD & 1:SNIM & 2:BIM & 3:RFGSM & 4:MIM & 5:DIM & 6:NIM & 7:VNIM & 8:VMIM & Clean\\
        \hline
        \multicolumn{2}{c|}{None-defense} & 0.002 & 0.000 & 0.000 & 0.000 & 0.000 & 0.000 & 0.001 & 0.000 & 0.000 & \textbf{0.898}\\
        \hline
        \multirow{5}{*}{\rotatebox{90}{AT}} & TRADES~\cite{TRADES} & 0.465 & 0.455 & 0.496 & 0.462 & 0.480 & 0.477 & 0.467 & 0.482 & 0.495 & 0.762\\
        & CLC~\cite{clc-defense} & 0.484 & 0.482 & 0.498 & 0.473 & 0.496 & 0.488 & 0.479 & 0.491 & 0.505 & 0.775\\
        & GAIRAT~\cite{GAIR} & 0.598 & 0.602 & 0.584 & 0.592 & 0.588 & 0.589 & 0.589 & 0.597 & 0.598 & 0.741\\
        & DMAT~\cite{DMAT} & 0.643 & 0.651 & 0.662 & 0.659 & 0.654 & 0.644 & 0.641 & 0.658 & 0.656 & 0.759 \\
        & MeanSparse~\cite{MeanSparse} & 0.694 & 0.693 & 0.705 & 0.701 & 0.713 & 0.718 & 0.700 & 0.716 & 0.714 & 0.788 \\
        \hline
        \multirow{2}{*}{\rotatebox{90}{AP}} & ADP~\cite{ADP} & 0.772 & 0.769 & 0.718 & 0.718 & 0.756 & 0.759 & 0.762 & 0.760 & 0.723 & 0.741 \\ 
        & LoRID~\cite{LoRID} & 0.783 & 0.780 & 0.752 & 0.756 & 0.773 & 0.776 & 0.775 & 0.779 & 0.748 & 0.740 \\ 
        \hline
        \multicolumn{2}{c|}{CAD(ours)} & \textbf{0.837} & \textbf{0.834} & \textbf{0.820} & \textbf{0.825} & \textbf{0.835} & \textbf{0.820} & \textbf{0.819} & \textbf{0.822} & \textbf{0.819} & {0.881} \\
        \rowcolor{lightgray} \multicolumn{2}{c|}{\textbf{$\mathrm{\textbf{CAD}}^\dagger$(ours)}} & 0.837 & 0.839 & 0.840 & 0.826 & 0.843 & 0.838 & 0.842 & 0.822 & 0.820 & 0.878 \\
        \bottomrule
    \end{tabular}
    \label{tab:imagenet100-main}
\end{table*}

\begin{table*}[ht]
\footnotesize
    \centering
    \tabcolsep=0.1cm
    \caption{Average incremental accuracy (AIAcc) against $\ell_{p = 0, 1, 2, \infty}$ attacks at each stage on CIFAR-10. DeepF. and SparseF. denote DeepFool and SparseFool respectively. $\mathrm{\textbf{CAD}}^\dagger$ is Premium-CAD without memory or defense budget constraints, serving as the empirical upper bound (gray background).
    }
    \begin{tabular}{c|l|ccccccccc}
        \toprule
        \multicolumn{2}{c|}{Method} & 0:PGD & 1:SNIM & 2:VNIM & 3:CW & 4:DeepF. & 5:EAD & 6:EADEN & 7:OnePixel & 8:SparseF.\\
        \hline
        \multicolumn{2}{c|}{None-defense} & 0.006 & 0.013 & 0.00 & 0.00 & 0.086 & 0.000 & 0.002 & 0.391 & 0.297 \\
        \hline
        \multirow{3}{*}{\rotatebox{90}{CL}} & EWC~\cite{EWC} & \textbf{0.955} & 0.682 & 0.670 & 0.237 & 0.236 & 0.237 & 0.269 & 0.259 & 0.222\\
        & SSRE~\cite{SSRE} & 0.891 & 0.825 & 0.781 & 0.658 & 0.584 & 0.506 & 0.475 & 0.424 & 0.378 \\ 
        & BEEF~\cite{BEEF} & 0.949 & 0.831 & 0.847 & 0.823 & 0.779 & 0.742 & 0.738 & 0.726 & 0.645 \\
        \hline
        \multirow{3}{*}{\rotatebox{90}{FCL}} & FACT~\cite{FACT} & 0.945 & 0.872 & 0.864 & 0.815 & 0.809 & 0.786 & 0.784 & 0.782 & 0.746 \\ 
        & OrCo~\cite{OrCo} & 0.946 & 0.871 & 0.865 & 0.821 & 0.815 & 0.802 & 0.793 & 0.789 & 0.752 \\
        & OrCo+ADBS~\cite{ADBS} & 0.946 & 0.872 & 0.866 & 0.823 & 0.818 & 0.803 & 0.795 & 0.790 & 0.753 \\
        \hline
        \multicolumn{2}{c|}{CAD(ours)} & 0.945 & \textbf{0.891} & \textbf{0.905} & \textbf{0.841} & \textbf{0.848} & \textbf{0.811} & \textbf{0.821} & \textbf{0.828} & \textbf{0.782}\\
        \rowcolor{lightgray} \multicolumn{2}{c|}{\textbf{$\mathrm{\textbf{CAD}}^\dagger$(ours)}} & 0.945 & 0.955 & 0.951 & 0.952 & 0.949 & 0.953 & 0.946 & 0.960 & 0.952\\
        \bottomrule
    \end{tabular}
    \label{tab:cifar10-CIL}
\end{table*}

\subsubsection{Defense Baselines}
We compare CAD with several state-of-the-art defense methods in terms of defense performance, including adversarial training approaches—TRADES~\cite{TRADES}, CLC~\cite{clc-defense}, GAIRAT~\cite{GAIR}, DMAT~\cite{DMAT}, and MeanSparse~\cite{MeanSparse}—as well as purification-based methods—ADP~\cite{ADP} and LoRID~\cite{LoRID}.
For adversarial training methods, gray-box attacks are not applicable during evaluation, as the defense model itself serves as the attack target.
Therefore, to highlight the superiority of CAD under gray-box attacks, we report the performance of adversarial training methods under black-box attacks in the same table for comparison.

To assess CAD's ability to mitigate catastrophic forgetting, we compare it with continual learning (CL) methods EWC~\cite{EWC}, SSRE~\cite{SSRE}, and BEEF~\cite{BEEF}, as well as few-shot continual learning (FCL) methods FACT~\cite{FACT}, OrCo~\cite{OrCo}, and ADBS~\cite{ADBS}, under a gray-box setting.


Additionally, we propose a variant of CAD, premium-CAD, as an empirical upper bound without memory or defense budget constraints.
In premium-CAD, attack-specific defense models are trained using an abundant defense budget and ensembled with the target model via a weight estimator.
Implementation details of both the baselines and premium-CAD are provided in the supplementary materials.

\subsubsection{Implementation Details}
Both the target and defense models use ResNet~\cite{ResNet} as the backbone: WideResNet-28-10 for CIFAR-10 following~\cite{JEM} and ResNet-50 for ImageNet-100 following~\cite{PAT}.
A 4-layer ConvNet is used as the weight estimator model.
The cosine classifier~\cite{cosine_classifier} is adopted as the classifier of the defense model.
The defense budget per class is set to $K=10$, with $T=8$ stages since there are 9 attacks (PGD as the initial attack). 
Following~\cite{FACT}, Beta distribution parameters are $\alpha = \beta = 2$, and trade-off parameters are $\gamma=0.01$.
The scaling hyperparameter of prototype augmentation is set to $\lambda=0.5$. 
Epochs for training and fine-tuning are set to $100$ and $4$, respectively.

\subsection{Comparisons to Baselines}
\label{exp:comparison}


To evaluate both adaptation and defense performance, we first compare CAD with baseline defense methods against nine adversarial attacks.
For CAD, the defense environment is dynamic, with attacks occurring incrementally at different stages.
In contrast, the baseline methods operate in a static defense environment where the order of attacks is not considered.
As shown in Table~\ref{tab:imagenet100-main}, despite the limited defense budget, CAD exhibits strong adaptability to diverse adversarial attacks.
On ImageNet-100, CAD outperforms MeanSparse and LoRID by $0.119$ and $0.057$ in average Acc, respectively.
These results highlight CAD’s capability for continuous adaptation to emerging threats.
Another notable advantage of CAD is its ability to maintain high Acc on clean images, achieving $0.881$—$0.093$ higher than MeanSparse and $0.140$ higher than ADP—approaching the target model's accuracy of $0.898$.
This improvement is attributed to the generic representations learned by the weight estimator from adversarial examples, which enable it to effectively distinguish between clean and adversarial inputs.

\begin{figure}[t]
    \centering
    \includegraphics[width=1\columnwidth]{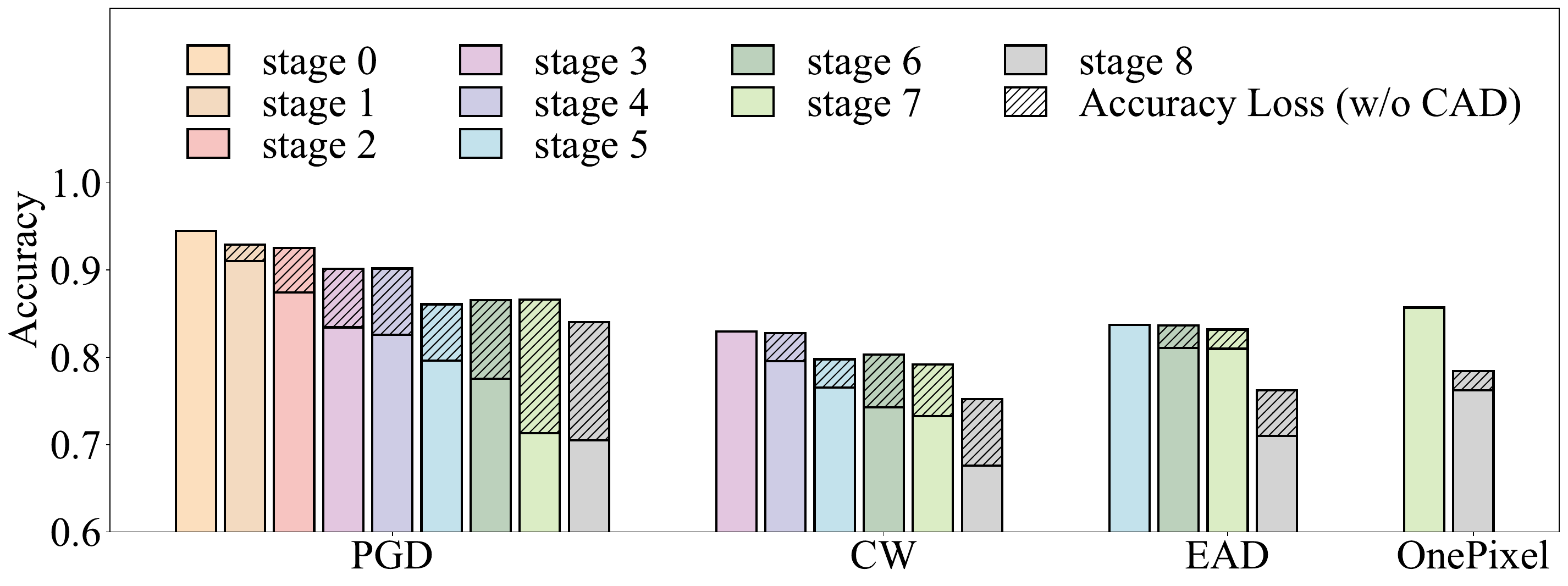}
    \caption{
    Verification of catastrophic forgetting in CAD and a vanilla model under the attack sequence from Table~\ref{tab:cifar10-CIL} on CIFAR-10.
    }
    \label{fig: forgetting}
\end{figure}

To evaluate CAD's ability to mitigate catastrophic forgetting, we compare it with CL and FCL methods on CIFAR-10.
As shown in Table~\ref{tab:cifar10-CIL}, CAD maintains robust defense performance despite varying attack types.
However, as the diversity of attacks increases, its performance gradually declines from $0.945$ to $0.782$.
This suggests that defending against an expanding array of attack types remains a significant challenge for continual defense methods.
Compared to the top baselines, CAD achieves an average AIAcc that is $0.024$ higher than OrCo and $0.023$ higher than ADBS across all stages.
Furthermore, we compare CAD with a vanilla model that serves as an empirical lower bound.
As shown in Figure~\ref{fig: forgetting}, CAD significantly reduces forgetting—without it, error rates on PGD and CW exceed 50\% by stage 8.


It is noteworthy that Premium-CAD surpasses all baseline methods.
Given a sufficient defense budget and memory resources, Premium-CAD emerges as the superior choice.
For more comparisons to baselines, please refer to the supplementary materials.


\begin{table}[t]
\footnotesize
\centering
\tabcolsep=0.19cm
\caption{
AIAcc against adversarial attacks and Acc on clean images for ablation study on CIFAR-10. “w/o $\mathcal{L}0$”, “w/o $\mathcal{L}{\mathrm{proto}}$”, and “w/o $f_p$” denote CAD without embedding space reservation, prototype augmentation, and model ensemble, respectively.
}
\begin{tabular}{l|cccc|c}
    \toprule
    Ablation & 0:PGD & 1:CW & 2:EAD & 3:OnePixel & Clean\\
    \hline
    w/o $\mathcal{L}_{0}$ & 0.945 & 824 & 0.761 & 0.718 & 0.958  \\
    w/o $\mathcal{L}_{\mathrm{proto}}$ & 0.944 & 0.895 & 0.800 & 0.785 & 0.958 \\ 
    w/o $f_p$ & 0.945 & 0.899 & 0.866 & 0.853 & 0.559 \\ 
    \hline
    \textbf{CAD} & \textbf{0.945} & \textbf{0.899} & \textbf{0.866} & \textbf{0.853} & \textbf{0.958}\\
    \bottomrule
\end{tabular}
\label{tab:ablation}
\end{table}

\subsection{Ablation Study}
\label{exp:ablation}
As argued, embedding reservation facilitates smoother adaptation and reduces overfitting, prototype augmentation preserves decision boundaries from previous stages, and model ensembling ensures high classification accuracy for both clean and adversarial images.  
To validate the contribution of each component, we conduct an ablation study using four attacks: PGD ($\ell_{\infty}$), CW ($\ell_2$), EAD ($\ell_1$), and OnePixel ($\ell_0$).  
As shown in Table~\ref{tab:ablation}, removing embedding space reservation ($\mathcal{L}_{0}$ in Eq.~\ref{eq:space_reserve}) leads to a significant drop in AIAcc at each stage due to the absence of a mechanism tailored for few-shot scenarios. In stage-3, AIAcc drops to $0.718$, which is $0.135$ lower than the original. 
When prototype augmentation ($\mathcal{L}_{\mathrm{proto}}$ in Eq.~\ref{eq:loss proto}) is removed, CAD loses its ability to prevent catastrophic forgetting, resulting in a noticeable decrease in AIAcc after stage-1, reaching $0.785$ in stage-3—$0.068$ lower than the original. 
Additionally, removing model ensembling ($f_p$ in Eq.~\ref{eq:model_ensmeble}) severely impairs CAD’s classification performance on clean images, reducing accuracy by $0.399$ compared to the original.
These results confirm the effectiveness of each component.
For analysis on the principles, ablation, and parameters, please refer to the supplementary materials.

\begin{table}[t]
\footnotesize
\centering
\tabcolsep=0.115cm
\caption{
Acc against unseen attacks after each adaptation on CIFAR-10.
}
\begin{tabular}{l|cccccccc}
    \toprule
    Attack (unseen) & 0:PGD & 1:SNIM & 2:BIM & 3:RFGSM & 4:MIM\\ 
    \hline
    DIM & 0.728 & 0.739 & 0.754 & 0.767 & 0.775\\ 
    NIM & 0.725 & 0.736 & 0.753 & 0.768 & 0.773\\ 
    VNIM & 0.708 & 0.724 & 0.745 & 0.752 & 0.764\\ 
    VMIM & 0.716 & 0.726 & 0.749 & 0.751 & 0.769\\ 
    \bottomrule
\end{tabular}
\label{tab:unseen}
\end{table}

\begin{table}[t]
\footnotesize
\centering
\tabcolsep=0.235cm
\caption{
Acc against unseen attacks after each adaptation on CIFAR-10 with a simulated attack sequence.
}
\begin{tabular}{l|cccccccc}
    \toprule
    Attack (unseen) & 0 & 10 & 20 & 30 & 40\\ 
    \hline
    DIM & 0.728 & 0.805 & 0.870 & 0.893 & 0.890\\ 
    NIM & 0.725 & 0.808 & 0.863 & 0.892 & 0.894\\ 
    VNIM & 0.708 & 0.796 & 0.855 & 0.879 & 0.884\\ 
    VMIM & 0.716 & 0.798 & 0.854 & 0.883 & 0.882\\ 
    \bottomrule
\end{tabular}
\label{tab:saturate}
\end{table}

\subsection{Defense against Unseen Attack and Saturation}

We also evaluated CAD’s performance on unseen attacks.
As shown in Table~\ref{tab:unseen}, adaptation to known attacks enables CAD to generalize to unknown ones, owing to the intrinsic similarities among attacks under the same norm~\cite{Wang:23selfperturbation}. 
With more adaptation stages, CAD learns diverse perturbation patterns, improving accuracy on unseen attacks from 0.719 to 0.770 after 4 stages. 
Eventually, when CAD has adapted to a sufficiently diverse set of attacks, the defense may reach a point of saturation. 
However, due to the current limited availability of attack types, this saturation point cannot yet be empirically determined.

To verify the saturation nature, we constructed simulated attacks using perturbation forgery~\cite{PFD} and tested CAD’s generalization ability after adapting to these synthetic attacks. 
As shown in Table~\ref{tab:saturate}, after 20 adaptation stages, CAD achieved an average classification accuracy of over 0.861 on unseen attacks. 
After 30 stages, the average accuracy on unseen attacks reached 0.887 and saturated—indicating that further adaptation did not yield additional improvements. This surpasses all baseline methods.

As mentioned earlier, CAD requires pre-allocated virtual classes for future classifier expansion. If the number of pre-allocated is insufficient, CAD’s scalability may be called into question. 
However, the saturation behavior suggests that with sufficient pre-allocated classes, CAD can reach a stage where it can defend against unseen attacks without further adaptation—alleviating concerns about its scalability.

\section{Conclusion}
We propose the first Continual Adversarial Defense (CAD) framework to address the dynamic nature of real-world attack environments, where no single defense can handle all threats. 
CAD is engineered to collect a defense budget from the Internet and dynamically adapt to a diverse range of attacks as they emerge in sequential stages.
In consideration of practical applicability, we formulate four principles: continual adaptation without catastrophic forgetting, few-shot adaptation, memory-efficient adaptation, and high classification accuracy on both clean and adversarial data. 
Extensive experiments demonstrate that CAD effectively adheres to the four principles and consistently outperforms baseline methods across multi-stage attacks.
Notably, its performance saturates as the number of attacks increases, highlighting its potential as a long-term defense solution.


\bibliography{advCIL}

\begin{thebibliography}{49}
\providecommand{\natexlab}[1]{#1}

\bibitem[{Ahmed, Kukleva, and Schiele(2024)}]{OrCo}
Ahmed, N.; Kukleva, A.; and Schiele, B. 2024.
\newblock Orco: Towards better generalization via orthogonality and contrast
  for few-shot class-incremental learning.
\newblock In \emph{Proceedings of the IEEE/CVF Conference on Computer Vision
  and Pattern Recognition}, 28762--28771.

\bibitem[{Amini et~al.(2024)Amini, Teymoorianfard, Ma, and
  Houmansadr}]{MeanSparse}
Amini, S.; Teymoorianfard, M.; Ma, S.; and Houmansadr, A. 2024.
\newblock MeanSparse: Post-training robustness enhancement through
  mean-centered feature sparsification.
\newblock \emph{arXiv preprint arXiv:2406.05927}.

\bibitem[{Chen, Li, and Zhang(2021)}]{clc-defense}
Chen, Z.; Li, Q.; and Zhang, Z. 2021.
\newblock Towards Robust Neural Networks via Close-loop Control.
\newblock In \emph{International Conference on Learning Representations}.

\bibitem[{Croce et~al.(2022)Croce, Gowal, Brunner, Shelhamer, Hein, and
  Cemgil}]{Test-time-defense}
Croce, F.; Gowal, S.; Brunner, T.; Shelhamer, E.; Hein, M.; and Cemgil, T.
  2022.
\newblock Evaluating the adversarial robustness of adaptive test-time defenses.
\newblock In \emph{International Conference on Machine Learning}, 4421--4435.
  PMLR.

\bibitem[{Deng et~al.(2009)Deng, Dong, Socher, Li, Li, and Fei-Fei}]{ImageNet}
Deng, J.; Dong, W.; Socher, R.; Li, L.-J.; Li, K.; and Fei-Fei, L. 2009.
\newblock ImageNet: A large-scale hierarchical image database.
\newblock In \emph{CVPR}.

\bibitem[{Dong and Xu(2023)}]{RPF}
Dong, M.; and Xu, C. 2023.
\newblock Adversarial Robustness via Random Projection Filters.
\newblock In \emph{CVPR}, 4077--4086.

\bibitem[{Dong et~al.(2018)Dong, Liao, Pang, Su, Zhu, Hu, and Li}]{MIFGSM}
Dong, Y.; Liao, F.; Pang, T.; Su, H.; Zhu, J.; Hu, X.; and Li, J. 2018.
\newblock Boosting Adversarial Attacks with Momentum.
\newblock In \emph{CVPR}, 9185--9193.

\bibitem[{Gidaris and Komodakis(2018)}]{cosine_classifier}
Gidaris, S.; and Komodakis, N. 2018.
\newblock Dynamic few-shot visual learning without forgetting.
\newblock In \emph{CVPR}, 4367--4375.

\bibitem[{Grathwohl et~al.(2020)Grathwohl, Wang, Jacobsen, Duvenaud, Norouzi,
  and Swersky}]{JEM}
Grathwohl, W.; Wang, K.-C.; Jacobsen, J.-H.; Duvenaud, D.; Norouzi, M.; and
  Swersky, K. 2020.
\newblock Your Classifier is Secretly an Energy Based Model and You Should
  Treat it Like One.
\newblock In \emph{ICLR}.

\bibitem[{He et~al.(2016)He, Zhang, Ren, and Sun}]{ResNet}
He, K.; Zhang, X.; Ren, S.; and Sun, J. 2016.
\newblock Deep residual learning for image recognition.
\newblock In \emph{CVPR}, 770--778.

\bibitem[{Hill, Mitchell, and Zhu(2020)}]{EBM}
Hill, M.; Mitchell, J.~C.; and Zhu, S.-C. 2020.
\newblock Stochastic Security: Adversarial Defense Using Long-Run Dynamics of
  Energy-Based Models.
\newblock In \emph{ICLR}.

\bibitem[{Jiang et~al.(2023)Jiang, Liu, Huang, Salzmann, and
  Susstrunk}]{FastAdv}
Jiang, Y.; Liu, C.; Huang, Z.; Salzmann, M.; and Susstrunk, S. 2023.
\newblock Towards Stable and Efficient Adversarial Training against $ l\_1 $
  Bounded Adversarial Attacks.
\newblock In \emph{ICML}, 15089--15104. PMLR.

\bibitem[{Kang et~al.(2021)Kang, Song, Ding, and Tay}]{lyapunov-defense}
Kang, Q.; Song, Y.; Ding, Q.; and Tay, W.~P. 2021.
\newblock Stable neural ode with lyapunov-stable equilibrium points for
  defending against adversarial attacks.
\newblock \emph{Advances in Neural Information Processing Systems}, 34:
  14925--14937.

\bibitem[{Kirkpatrick et~al.(2017)Kirkpatrick, Pascanu, Rabinowitz, Veness,
  Desjardins, Rusu, Milan, Quan, Ramalho, Grabska-Barwinska et~al.}]{EWC}
Kirkpatrick, J.; Pascanu, R.; Rabinowitz, N.; Veness, J.; Desjardins, G.; Rusu,
  A.~A.; Milan, K.; Quan, J.; Ramalho, T.; Grabska-Barwinska, A.; et~al. 2017.
\newblock Overcoming catastrophic forgetting in neural networks.
\newblock \emph{Proceedings of the national academy of sciences}, 114(13):
  3521--3526.

\bibitem[{Krizhevsky, Hinton et~al.(2009)}]{CIFAR10}
Krizhevsky, A.; Hinton, G.; et~al. 2009.
\newblock Learning multiple layers of features from tiny images.
\newblock \emph{Technical report}.

\bibitem[{Kurakin, Goodfellow, and Bengio(2016)}]{BIM}
Kurakin, A.; Goodfellow, I.; and Bengio, S. 2016.
\newblock Adversarial Machine Learning at Scale.
\newblock arXiv:1611.01236.

\bibitem[{Laidlaw, Singla, and Feizi(2020)}]{PAT}
Laidlaw, C.; Singla, S.; and Feizi, S. 2020.
\newblock Perceptual Adversarial Robustness: Defense Against Unseen Threat
  Models.
\newblock In \emph{ICLR}.

\bibitem[{Li et~al.(2025)Li, Tan, Yang, Cheng, Dong, and Yang}]{ADBS}
Li, L.; Tan, Y.; Yang, S.; Cheng, H.; Dong, Y.; and Yang, L. 2025.
\newblock Adaptive Decision Boundary for Few-Shot Class-Incremental Learning.
\newblock In \emph{Proceedings of the AAAI Conference on Artificial
  Intelligence}, volume~39, 18359--18367.

\bibitem[{Li, Xin, and Liu(2022)}]{ASODE-dynamic}
Li, X.; Xin, Z.; and Liu, W. 2022.
\newblock Defending against adversarial attacks via neural dynamic system.
\newblock \emph{Advances in Neural Information Processing Systems}, 35:
  6372--6383.

\bibitem[{Li et~al.(2023)Li, Xu, Su, and Jia}]{li2023robustness}
Li, Y.; Xu, X.; Su, Y.; and Jia, K. 2023.
\newblock On the robustness of open-world test-time training: Self-training
  with dynamic prototype expansion.
\newblock In \emph{Proceedings of the IEEE/CVF International Conference on
  Computer Vision}, 11836--11846.

\bibitem[{Lin et~al.(2019)Lin, Song, He, Wang, and Hopcroft}]{NIFGSM}
Lin, J.; Song, C.; He, K.; Wang, L.; and Hopcroft, J.~E. 2019.
\newblock Nesterov Accelerated Gradient and Scale Invariance for Adversarial
  Attacks.
\newblock In \emph{ICLR}.

\bibitem[{Liu et~al.(2022)Liu, Li, Schiele, and Sun}]{lyy2022online}
Liu, Y.; Li, Y.; Schiele, B.; and Sun, Q. 2022.
\newblock Online Hyperparameter Optimization for Class-Incremental Learning.
\newblock In \emph{AAAI}.

\bibitem[{Madry et~al.(2017)Madry, Makelov, Schmidt, Tsipras, and Vladu}]{PGD}
Madry, A.; Makelov, A.; Schmidt, L.; Tsipras, D.; and Vladu, A. 2017.
\newblock Towards Deep Learning Models Resistant to Adversarial Attacks.
\newblock arXiv:1706.06083.

\bibitem[{Nie et~al.(2022)Nie, Guo, Huang, Xiao, Vahdat, and
  Anandkumar}]{DiffPure}
Nie, W.; Guo, B.; Huang, Y.; Xiao, C.; Vahdat, A.; and Anandkumar, A. 2022.
\newblock Diffusion Models for Adversarial Purification.
\newblock In \emph{ICML}.

\bibitem[{Pei et~al.(2025)Pei, Lyu, Chen, Ma, Xu, Sun, and Huang}]{DnC}
Pei, G.; Lyu, S.; Chen, G.; Ma, K.; Xu, Q.; Sun, Y.; and Huang, Q. 2025.
\newblock Divide and conquer: Heterogeneous noise integration for
  diffusion-based adversarial purification.
\newblock In \emph{Proceedings of the Computer Vision and Pattern Recognition
  Conference}, 29268--29277.

\bibitem[{Rebuffi et~al.(2017)Rebuffi, Kolesnikov, Sperl, and Lampert}]{iCarl}
Rebuffi, S.-A.; Kolesnikov, A.; Sperl, G.; and Lampert, C.~H. 2017.
\newblock icarl: Incremental classifier and representation learning.
\newblock In \emph{CVPR}, 2001--2010.

\bibitem[{Snell, Swersky, and Zemel(2017)}]{protonet}
Snell, J.; Swersky, K.; and Zemel, R. 2017.
\newblock Prototypical networks for few-shot learning.
\newblock \emph{NeurIPS}, 30.

\bibitem[{Taran et~al.(2019)Taran, Rezaeifar, Holotyak, and
  Voloshynovskiy}]{Taran:19defending}
Taran, O.; Rezaeifar, S.; Holotyak, T.; and Voloshynovskiy, S. 2019.
\newblock Defending against adversarial attacks by randomized diversification.
\newblock In \emph{CVPR}, 11226--11233.

\bibitem[{Tian, Krishnan, and Isola(2020)}]{ImageNet100}
Tian, Y.; Krishnan, D.; and Isola, P. 2020.
\newblock Contrastive multiview coding.
\newblock In \emph{Computer Vision--ECCV 2020: 16th European Conference,
  Glasgow, UK, August 23--28, 2020, Proceedings, Part XI 16}, 776--794.
  Springer.

\bibitem[{Tram{\`e}r et~al.(2018)Tram{\`e}r, Kurakin, Papernot, Goodfellow,
  Boneh, and McDaniel}]{Ensemble_AT}
Tram{\`e}r, F.; Kurakin, A.; Papernot, N.; Goodfellow, I.; Boneh, D.; and
  McDaniel, P. 2018.
\newblock Ensemble Adversarial Training: Attacks and Defenses.
\newblock In \emph{ICLR}.

\bibitem[{Verma et~al.(2019)Verma, Lamb, Beckham, Najafi, Mitliagkas,
  Lopez-Paz, and Bengio}]{manifoldmix}
Verma, V.; Lamb, A.; Beckham, C.; Najafi, A.; Mitliagkas, I.; Lopez-Paz, D.;
  and Bengio, Y. 2019.
\newblock Manifold mixup: Better representations by interpolating hidden
  states.
\newblock In \emph{ICML}, 6438--6447.

\bibitem[{Wang et~al.(2022)Wang, Zhou, Liu, Ye, Bian, Zhan, and Zhao}]{BEEF}
Wang, F.-Y.; Zhou, D.-W.; Liu, L.; Ye, H.-J.; Bian, Y.; Zhan, D.-C.; and Zhao,
  P. 2022.
\newblock BEEF: Bi-compatible class-incremental learning via energy-based
  expansion and fusion.
\newblock In \emph{ICLR}.

\bibitem[{Wang et~al.(2024)Wang, Li, Luo, Ling, Huang, Jia, and Yu}]{PFD}
Wang, Q.; Li, C.; Luo, Y.; Ling, H.; Huang, S.; Jia, R.; and Yu, N. 2024.
\newblock Detecting Adversarial Data using Perturbation Forgery.
\newblock \emph{arXiv preprint arXiv:2405.16226}.

\bibitem[{Wang et~al.(2023{\natexlab{a}})Wang, Xian, Ling, Zhang, Lin, Li,
  Chen, and Yu}]{Wang:23selfperturbation}
Wang, Q.; Xian, Y.; Ling, H.; Zhang, J.; Lin, X.; Li, P.; Chen, J.; and Yu, N.
  2023{\natexlab{a}}.
\newblock Detecting Adversarial Faces Using Only Real Face Self-Perturbations.
\newblock In \emph{IJCAI}, 1488--1496.
\newblock Main Track.

\bibitem[{Wang and He(2021)}]{VNIFGSM}
Wang, X.; and He, K. 2021.
\newblock Enhancing the transferability of adversarial attacks through variance
  tuning.
\newblock In \emph{CVPR}, 1924--1933.

\bibitem[{Wang et~al.(2023{\natexlab{b}})Wang, Pang, Du, Lin, Liu, and
  Yan}]{DMAT}
Wang, Z.; Pang, T.; Du, C.; Lin, M.; Liu, W.; and Yan, S. 2023{\natexlab{b}}.
\newblock Better Diffusion Models Further Improve Adversarial Training.
\newblock In \emph{ICML}.

\bibitem[{Wu et~al.(2021)Wu, Su, Lyu, and King}]{DIFGSM}
Wu, W.; Su, Y.; Lyu, M.~R.; and King, I. 2021.
\newblock Improving the Transferability of Adversarial Samples with Adversarial
  Transformations.
\newblock In \emph{CVPR}, 9020--9029.

\bibitem[{Yin et~al.(2024)Yin, Yao, Xiao, and Long}]{ea-dynamic}
Yin, S.; Yao, K.; Xiao, Z.; and Long, J. 2024.
\newblock Embracing Adaptation: An Effective Dynamic Defense Strategy Against
  Adversarial Examples.
\newblock In \emph{Proceedings of the 32nd ACM International Conference on
  Multimedia}, 9435--9444.

\bibitem[{Yoon, Hwang, and Lee(2021)}]{ADP}
Yoon, J.; Hwang, S.~J.; and Lee, J. 2021.
\newblock Adversarial purification with score-based generative models.
\newblock In \emph{ICML}, 12062--12072.

\bibitem[{Zhang et~al.(2021)Zhang, Song, Lin, Zheng, Pan, and Xu}]{CEC}
Zhang, C.; Song, N.; Lin, G.; Zheng, Y.; Pan, P.; and Xu, Y. 2021.
\newblock Few-shot incremental learning with continually evolved classifiers.
\newblock In \emph{CVPR}, 12455--12464.

\bibitem[{Zhang et~al.(2019)Zhang, Yu, Jiao, Xing, El~Ghaoui, and
  Jordan}]{TRADES}
Zhang, H.; Yu, Y.; Jiao, J.; Xing, E.; El~Ghaoui, L.; and Jordan, M. 2019.
\newblock Theoretically principled trade-off between robustness and accuracy.
\newblock In \emph{ICML}, 7472--7482.

\bibitem[{Zhang et~al.(2020)Zhang, Zhu, Niu, Han, Sugiyama, and
  Kankanhalli}]{GAIR}
Zhang, J.; Zhu, J.; Niu, G.; Han, B.; Sugiyama, M.; and Kankanhalli, M. 2020.
\newblock Geometry-aware Instance-reweighted Adversarial Training.
\newblock In \emph{ICLR}.

\bibitem[{Zhang and Sabuncu(2018)}]{CrossEntropyLoss}
Zhang, Z.; and Sabuncu, M. 2018.
\newblock Generalized Cross Entropy Loss for Training Deep Neural Networks with
  Noisy Labels.
\newblock In Bengio, S.; Wallach, H.; Larochelle, H.; Grauman, K.;
  Cesa-Bianchi, N.; and Garnett, R., eds., \emph{NeurIPS}, volume~31.

\bibitem[{Zhou et~al.(2022)Zhou, Wang, Ye, Ma, Pu, and Zhan}]{FACT}
Zhou, D.-W.; Wang, F.-Y.; Ye, H.-J.; Ma, L.; Pu, S.; and Zhan, D.-C. 2022.
\newblock Forward compatible few-shot class-incremental learning.
\newblock In \emph{CVPR}, 9046--9056.

\bibitem[{Zhou and Hua(2024)}]{AIR-CAD}
Zhou, Y.; and Hua, Z. 2024.
\newblock Defense without Forgetting: Continual Adversarial Defense with
  Anisotropic \& Isotropic Pseudo Replay.
\newblock In \emph{Proceedings of the IEEE/CVF Conference on Computer Vision
  and Pattern Recognition}, 24263--24272.

\bibitem[{Zhu et~al.(2021{\natexlab{a}})Zhu, Cheng, Zhang, and
  Liu}]{Dual_Aug_CIL}
Zhu, F.; Cheng, Z.; Zhang, X.-y.; and Liu, C.-l. 2021{\natexlab{a}}.
\newblock Class-incremental learning via dual augmentation.
\newblock In \emph{NeurIPS}, volume~34, 14306--14318.

\bibitem[{Zhu et~al.(2021{\natexlab{b}})Zhu, Zhang, Wang, Yin, and Liu}]{PASS}
Zhu, F.; Zhang, X.-Y.; Wang, C.; Yin, F.; and Liu, C.-L. 2021{\natexlab{b}}.
\newblock Prototype augmentation and self-supervision for incremental learning.
\newblock In \emph{CVPR}, 5871--5880.

\bibitem[{Zhu et~al.(2022)Zhu, Zhai, Cao, Luo, and Zha}]{SSRE}
Zhu, K.; Zhai, W.; Cao, Y.; Luo, J.; and Zha, Z.-J. 2022.
\newblock Self-sustaining representation expansion for non-exemplar
  class-incremental learning.
\newblock In \emph{CVPR}, 9296--9305.

\bibitem[{Zollicoffer et~al.(2025)Zollicoffer, Vu, Nebgen, Castorena,
  Alexandrov, and Bhattarai}]{LoRID}
Zollicoffer, G.; Vu, M.~N.; Nebgen, B.; Castorena, J.; Alexandrov, B.; and
  Bhattarai, M. 2025.
\newblock Lorid: Low-rank iterative diffusion for adversarial purification.
\newblock In \emph{Proceedings of the AAAI Conference on Artificial
  Intelligence}, volume~39, 23081--23089.

\end{thebibliography}


\end{document}